\documentclass{article}

\usepackage[preprint]{corl_2026}

\usepackage{comment}
\usepackage{booktabs}
\usepackage{tabularx}
\usepackage{graphicx}
\usepackage{float}
\usepackage{svg}
\usepackage{amsmath}
\usepackage{amssymb} 
\usepackage{algorithm}
\usepackage{multirow}
\usepackage{algpseudocode}

\usepackage{titlesec}

\title{AXIS: A Growable Community-Driven Data Engine for Scalable Robot Manipulation}

\author{%
\makebox[0pt][c]{%
\begin{minipage}{\textwidth}
\centering
\vspace*{0.35cm}
{\normalsize
\textbf{Mengfei Zhao}$^{*,\ddag}$,
\textbf{Dihong Huang}$^{1,2*}$,
\textbf{Yikai Tang}$^{2*}$,
\textbf{Peihao Li}$^{2*}$,
\textbf{Mingxuan Yan}$^{3*}$,\\
\textbf{Ruiqi Zhuang}$^{1*}$,
\textbf{Yanjia Huang}$^{4*}$,
\textbf{Jie Wang}$^{1,5}$,
\textbf{Hai Zhai}$^{1}$,
\textbf{Tony Zhou}$^{1,6}$,\\
\textbf{Rui Zhang}$^{1,7}$,
\textbf{Zhexi Luo}$^{1,8}$,
\textbf{Yuchen Huang}$^{1,7}$,
\textbf{Jianfei Yang}$^{9\ddag}$,
\textbf{Jiachen Li}$^{3\ddag}$%
}\\[0.6em]
{\small\normalfont\mdseries
$^{1}$Axis Robotics
\quad
$^{2}$University of California, Berkeley
\quad
$^{3}$Georgia Institute of Technology\\
$^{4}$Texas A\&M University
\quad
$^{5}$Johns Hopkins University
\quad
$^{6}$University of Pennsylvania\\
$^{7}$University of Michigan
\quad
$^{8}$National University of Singapore
\quad
$^{9}$Nanyang Technological University%
}\\[0.3em]
{\scriptsize\normalfont\mdseries
$^{*}$Equal contribution \quad
$^{\ddag}$\texttt{zhaomengfei248@gmail.com, jianfei.yang@ntu.edu.sg, jiachen\_li@gatech.edu}%
}
\end{minipage}%
}%
}

\begin{document}

\maketitle
\vspace{-1.2cm}

\begin{figure}[H]
    \centering
    \makebox[\textwidth][c]{%
        \includegraphics[width=1.\textwidth]{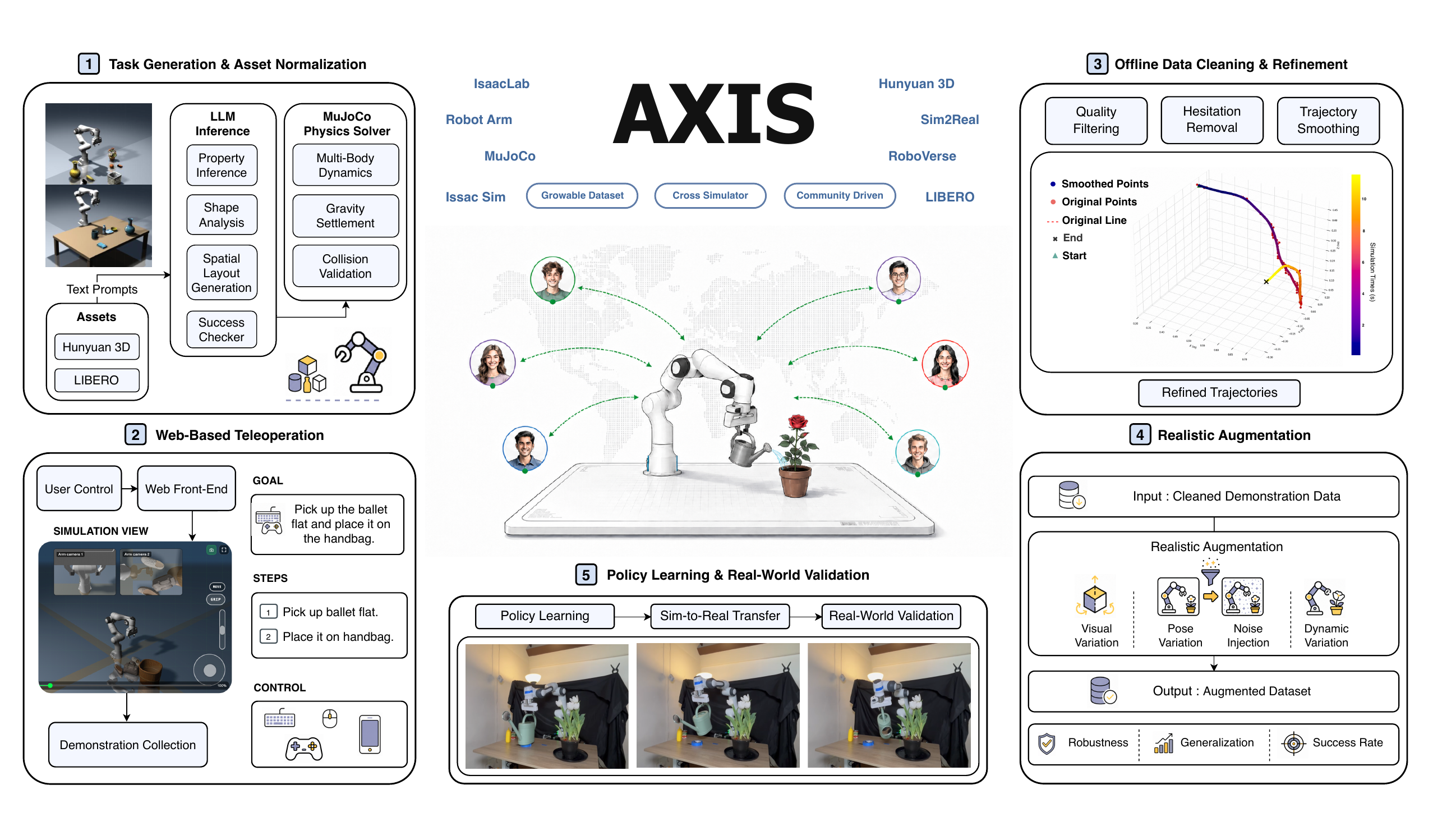}
    }
    \vspace{-0.8cm}
\caption{AXIS: A growable community-driven data engine unifying task generation, web teleoperation, data refinement, realistic augmentation, and policy learning with real-world validation.
}
    \label{fig:teaser}
    \vspace{-0.2cm}

\end{figure}


\begin{abstract}
Learning effective robot manipulation policies requires diverse, high-quality demonstrations, yet existing data pipelines are often difficult to scale because they rely on specialized hardware, centralized operators, or fixed task suites.
We present AXIS, a growable community-driven data engine and benchmark for scalable robot learning, which enables browser-based teleoperation for large-scale demonstration collection, automatically generates and validates new manipulation tasks, and transforms community-collected demonstrations into training-ready data through automated success checking, quality filtering, trajectory smoothing, and visual and physics-based augmentation.
The AXIS dataset currently contains 207 diverse tasks and 50K+ trajectories. Meanwhile, AXIS organizes data into task snapshots and evaluates policies with a systematic held-out protocol.
We compare vision-language-action (VLA) policies under a unified AXIS evaluation suite and analyze scaling behavior across different data volumes.
Continual pretraining on AXIS substantially improves the overall success rate of $\pi_{0.5}$ by 5.8\%, outperforms the model pretrained on RoboCasa365 by 37.3\%, and exhibits consistent scaling with increasing data volume, with the largest gains observed under layout, sensor-noise, and camera perturbations.

\textbf{Project website:} \url{https://axisaiorg.github.io/AXIS-V1/}
\end{abstract}



\section{Introduction}
\vspace{-0.2cm}

General-purpose robot manipulation requires policies that can act across diverse objects, scenes, and task goals. Recent progress in imitation learning and vision-language-action (VLA) models has shown that policy performance depends heavily on the scale, diversity, and quality of robot demonstration data~\cite{brohan2023rt1roboticstransformerrealworld, geng2025roboverse}. 
However, collecting physical robot demonstrations remains expensive, teleoperation systems often require specialized hardware or local simulator installations, and most public manipulation datasets remain fixed after a one-time collection effort~\cite{duan2022survey, choi2021use}. As a result, robot datasets have grown much more slowly than the models they are intended to train.
A fundamental limitation of existing manipulation data pipelines is that data collection remains largely closed and centralized. Expert operators or dedicated teams gather demonstrations on local hardware, process them offline, and release the resulting dataset as a fixed benchmark~\cite{choi2021use, esser2022guided}. While this paradigm offers strong control over data quality, it does not naturally support continual task expansion, broad community participation, or rapid iteration as robot capabilities evolve. Scaling robot learning instead requires data infrastructure that can continuously expand along multiple dimensions, including tasks, objects, scenes, contributors, visual conditions, and physical variations.

We argue that the next generation of robot manipulation datasets should be \emph{growable}: rather than serving as static collections of demonstrations, they should provide mechanisms for continuously generating, collecting, validating, processing, and evaluating new data. 
Such growth should also be structured, organizing new tasks and demonstrations into reproducible dataset snapshots that support consistent benchmarking over time.
To this end, we introduce \textbf{AXIS}, a community-driven data engine and benchmark for scalable robot manipulation. AXIS is built on a simple principle: make large-scale data collection broadly accessible while scaling high-quality data processing on demand. The platform currently targets tabletop manipulation with a Franka Research 3 robot equipped with a parallel-jaw gripper, supporting a wide range of tasks including pick-and-place, stacking, pushing, pouring, articulated-object manipulation, and tool-use behaviors. Each task is specified by a natural-language instruction, a parameterized scene configuration, and a structured success checker.

AXIS comprises three tightly integrated layers. The \emph{infrastructure layer} combines automated task generation with browser-based MuJoCo-WASM teleoperation. New tasks are generated from high-level instructions, object assets, scene layouts, and task-specific success conditions, then deployed through a shared web interface where users collect demonstrations using commodity input devices. The \emph{dataset layer} transforms raw community demonstrations into training-ready data through unified trajectory representation, automated success validation, quality filtering, static-segment removal, trajectory smoothing, resampling, and IsaacSim-based visual and physics augmentation. Finally, the \emph{model layer} supports training and evaluation of both conventional visuomotor imitation learning policies and modern VLA models using shared task definitions and success checkers.
AXIS is also designed to make dataset growth measurable. It organizes demonstrations into progressively larger task snapshots and evaluates policies using a fixed held-out protocol, which enables controlled scaling studies in which the policy architecture, training budget, rollout protocol, evaluation tasks, and success criteria remain fixed while only the training snapshot grows. 

\textbf{Contributions.}
First, we present AXIS, a web-based, community-driven infrastructure for scalable manipulation data collection through browser-based MuJoCo-WASM teleoperation and automated task generation. Second, we build a growable manipulation dataset comprising 207 tasks, 50K+ trajectories, together with a unified data processing pipeline for trajectory standardization, success validation, quality filtering, temporal smoothing, resampling, IsaacSim-based visual and physics augmentation, and versioned task snapshotting. Third, we introduce a systematic evaluation protocol for studying policy learning and within-embodiment dataset scaling, including standardized benchmark comparisons and controlled scaling experiments using the AXIS-25\%/50\%/100\% dataset snapshots under a fixed held-out evaluation suite.

\textbf{Key findings.}
We evaluate AXIS through two complementary experimental protocols. First, we compare representative imitation learning and VLA policies under a unified evaluation suite, using our dataset as a consistent benchmark. Second, we investigate dataset scaling by training a fixed policy recipe on progressively larger dataset snapshots while keeping the optimization budget and held-out evaluation protocol unchanged. The results show that large-scale, community-collected demonstrations, combined with automated processing and augmentation, substantially improve downstream policy learning. Continual pretraining of $\pi_{0.5}$ on AXIS-100\% improves the overall LIBERO-Plus success rate by 5.8\% and outperforms a volume-matched RoboCasa365 continual pretraining baseline by 37.3\%. Performance scales consistently with dataset size (84.7\%/85.7\%/88.8\% for AXIS-25\%/50\%/100\%), with the largest improvements observed under camera (+15.6\%), sensor-noise (+16.6\%), layout (+3.1\%), and robot-pose (+5.1\%) perturbations, demonstrating that the growable design of AXIS provides a practical path toward scalable robot manipulation.

\vspace{-0.1cm}
\section{Related Work}
\vspace{-0.1cm}

\noindent \textbf{Large-scale and crowdsourced robot manipulation data.}
Large-scale robot learning increasingly relies on broad manipulation corpora spanning multi-robot transfer, cross-environment trajectories, in-the-wild demonstrations, multi-embodiment data, and institutional aggregation~\cite{dasari2020robonetlargescalemultirobotlearning,ebert2021bridgedataboostinggeneralization,pmlr-v229-walke23a,khazatsky2025droidlargescaleinthewildrobot,wu2025robomind,10611477}. These datasets have supported scalable robot policies~\cite{brohan2023rt1roboticstransformerrealworld,pmlr-v229-zitkovich23a}, while recent benchmarks extend coverage to lifelong learning, household tasks, compositional reasoning, memory, grasping, and part-level supervision~\cite{NEURIPS2023_8c3c6668,robocasa2024,robocasa365,chen2025robohimanhierarchicalevaluationparadigm,dai2026robommebenchmarkingunderstandingmemory,srinivas2025graspfactorylargeobjectcentricgrasping,li2025rose,yin2025partinstruct}. Crowdsourced systems broaden collection~\cite{mandlekar2018roboturk,mirchandani2025robocade}, but many resources remain fixed releases or depend on physical robots and specific setups. AXIS instead uses browser-based simulated teleoperation for community collection and organizes data into versioned snapshots, making dataset growth measurable rather than a one-time release.

\noindent \textbf{Teleoperation interfaces for robot data collection.}
Robot demonstrations are commonly collected by teleoperation, ranging from kinesthetic teaching~\cite{kormushev2011positionalforce} and leader-follower rigs~\cite{wu2023gello,zhao2023aloha,zou2025uarm} to commodity devices~\cite{robocasa2024,honerkamp2025zerocost,zhou2022teleman,george2025openvr,lu2025know,wang2024eve} and wearable capture~\cite{wang2024dexcap,xu2025dexumi}. These interfaces trade off precision, cost, embodiment fidelity, calibration burden, and accessibility. Web systems reduce onboarding, but prior work often targets remote physical robots~\cite{mirchandani2025robocade,fok2016webbased}, leaving collection constrained by robot availability, maintenance, and safety; high-fidelity simulators meanwhile require heavier installation and GPU resources~\cite{nvidia2025isaacsim,xiang2020sapien}. AXIS separates collection from expensive simulation by using MuJoCo-WASM for browser-based teleoperation and reserving high-fidelity replay and augmentation for backend processing.

\noindent \textbf{Simulation, task generation, and synthetic augmentation.}
Simulation supports scalable task variation, replay, and validation, spanning lightweight engines for interaction and prototyping~\cite{todorov2012mujoco,koenig2004gazebo,rohmer2013coppeliasim,drake,coumans2016pybullet} and GPU-oriented platforms for parallel simulation, rendering, and complex scenes~\cite{makoviychuk2021isaacgym,zakka2025mujoco_playground,freeman2021brax,genesis2024engine,nvidia2025isaacsim,xiang2020sapien}. Standardized task suites define common interfaces and success criteria~\cite{gu2023maniskill2,tao2024maniskill3,zhu2020robosuite,robocasa365,geng2025roboverse}; data-generation methods expand demonstrations through trajectory retargeting and synthesis~\cite{mandlekar2023mimicgen,jiang2024dexmimicgen}; and LLM/VLM-driven systems scale task and asset generation~\cite{wang2024gensim,hua2024gensim2,wang2024robogen,he2026fishbone3dassetmillion}. AXIS unifies these components into a single growable data pipeline through automated task generation, browser-based teleoperation, and IsaacSim-based augmentation.

\noindent \textbf{VLA models and evaluation.}
Scaling data is closely tied to training and evaluating general manipulation policies. Diffusion-style policies and action-chunking transformers remain strong baselines, with ACT commonly used for chunked action prediction~\cite{zhao2023aloha}. VLA models finetune large vision-language backbones on robot trajectories and have advanced instruction-following manipulation~\cite{octo_2024,kim2024openvla,kim2025openvla_oft,liu2024rdt,li2024cogact,shi2025memoryvla,black2024pi0,intelligence2025pi05,nvidia2025groot,chen2025large}. Evaluation has likewise moved toward shared interfaces and robustness probes, including lifelong tasks, perturbation-aware extensions, sim-to-real-correlated evaluation, and combinatorial scene shifts~\cite{NEURIPS2023_8c3c6668,libero_plus_2025,li2024simplerenv,pumacay2024colosseum}. 
Rather than freezing a single suite, AXIS releases versioned training snapshots with a fixed held-out protocol so policies can be compared under shared data, observation, and success-checker conventions as data grows.

\section{Scalable Robot Data Collection}

The dataset layer converts raw community demonstrations into training-ready robot trajectories through validation, filtering, refinement, and augmentation. Since demonstrations are collected from human operators rather than scripted controllers, the resulting data contains behavioral diversity in approach direction, grasp timing, correction behavior, and execution style.

Completed demonstrations are serialized into a unified trajectory format. Each trajectory contains task metadata, robot embodiment, environment identifier, simulator version, timestamps, observations, robot states, actions, and success information. Optional fields include object states, camera streams, segmentation markers, and failure labels when available. This format makes demonstrations replayable, filterable, and compatible with downstream policy learning.

Raw community demonstrations may contain jitter, stalls, corrupted frames, discontinuities, or incorrect success labels. The dataset is therefore filtered using consistency checks for missing keys, anomalous numerical values, invalid state transitions, physically implausible frame-to-frame changes, and failed task completion. Although frontend dumps include success metadata, AXIS reuses task-specific success checkers during backend validation rather than relying solely on frontend success flags. Retained trajectories are then refined by removing static or corrupted segments, smoothing high-frequency teleoperation artifacts, and resampling to a fixed control frequency. This produces more stable action sequences while preserving the geometric structure of the demonstrated manipulation behavior.
Cleaned demonstrations are replayed in an IsaacSim backend for visual and physics augmentation. Physical randomization varies object poses, clutter configurations, mass, friction, and related dynamics parameters. Rendering randomization varies camera viewpoints, lighting, textures, and visual appearance. These perturbations are applied while preserving task semantics, expanding the training distribution without requiring additional human demonstrations for every variation. The result is a scalable pipeline that converts accessible web-collected demonstrations into higher-quality data for VLA and robot policy learning.

\vspace{-0.2cm}
\section{The AXIS Franka Dataset}
\vspace{-0.2cm}

\begin{figure}
    \centering
    \includegraphics[width=1\linewidth]{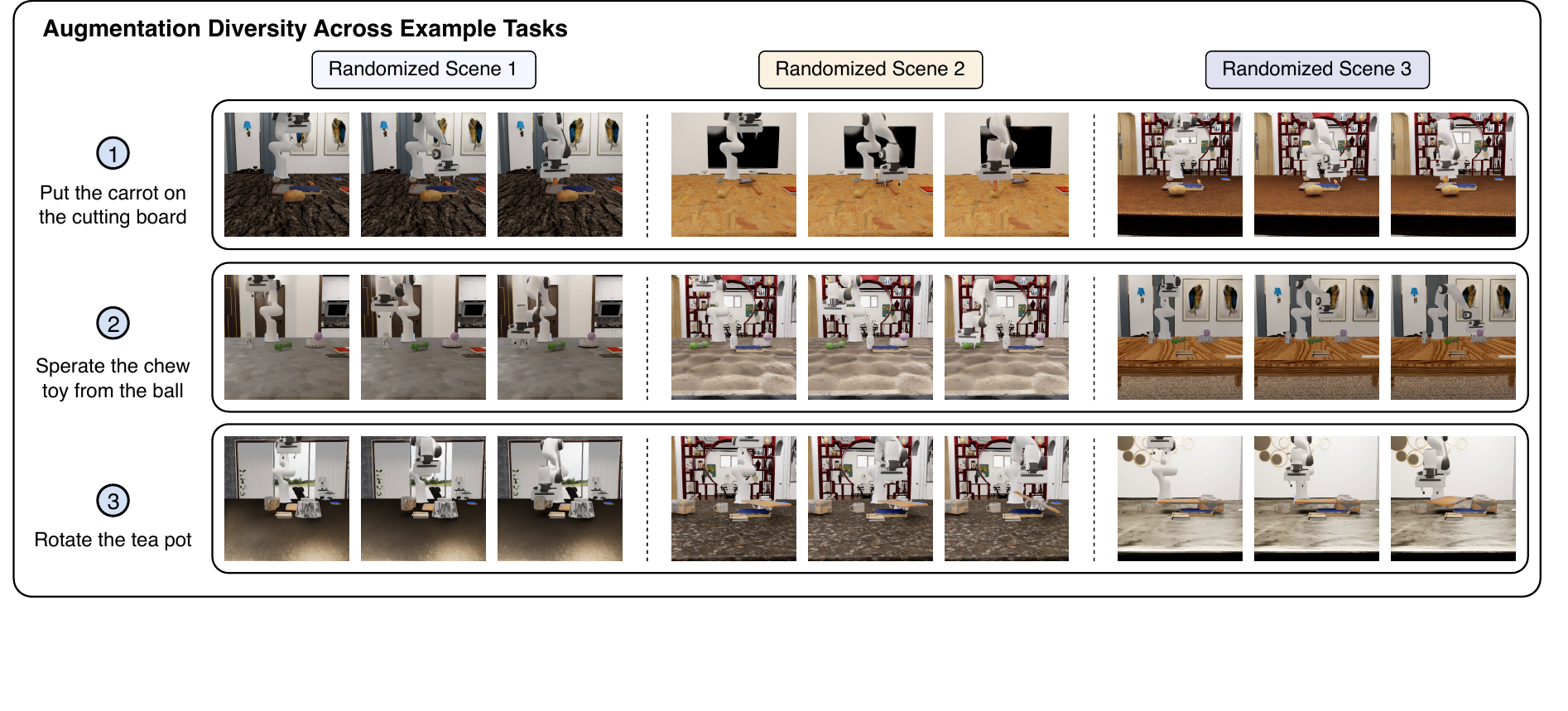}
    \vspace{-1.4cm}
    \caption{Visualization of augmentation diversity for three example tasks. For each task (row), we show three augmented variations (columns) obtained by randomizing appearance, lighting, environment, and initial object positions. Across augmentations, we vary materials and textures (e.g., table, floor, objects), lighting conditions (e.g., brightness and shadows), background and scene layout (e.g., walls, furniture, decorations), and initial states (e.g., object positions and orientations). This process substantially increases data diversity and improves policy robustness to environmental variations.}
    \vspace{-0.5cm}
    \label{fig:data_demo}
\end{figure}

\begin{figure}
    \centering
    \includegraphics[width=1\linewidth]{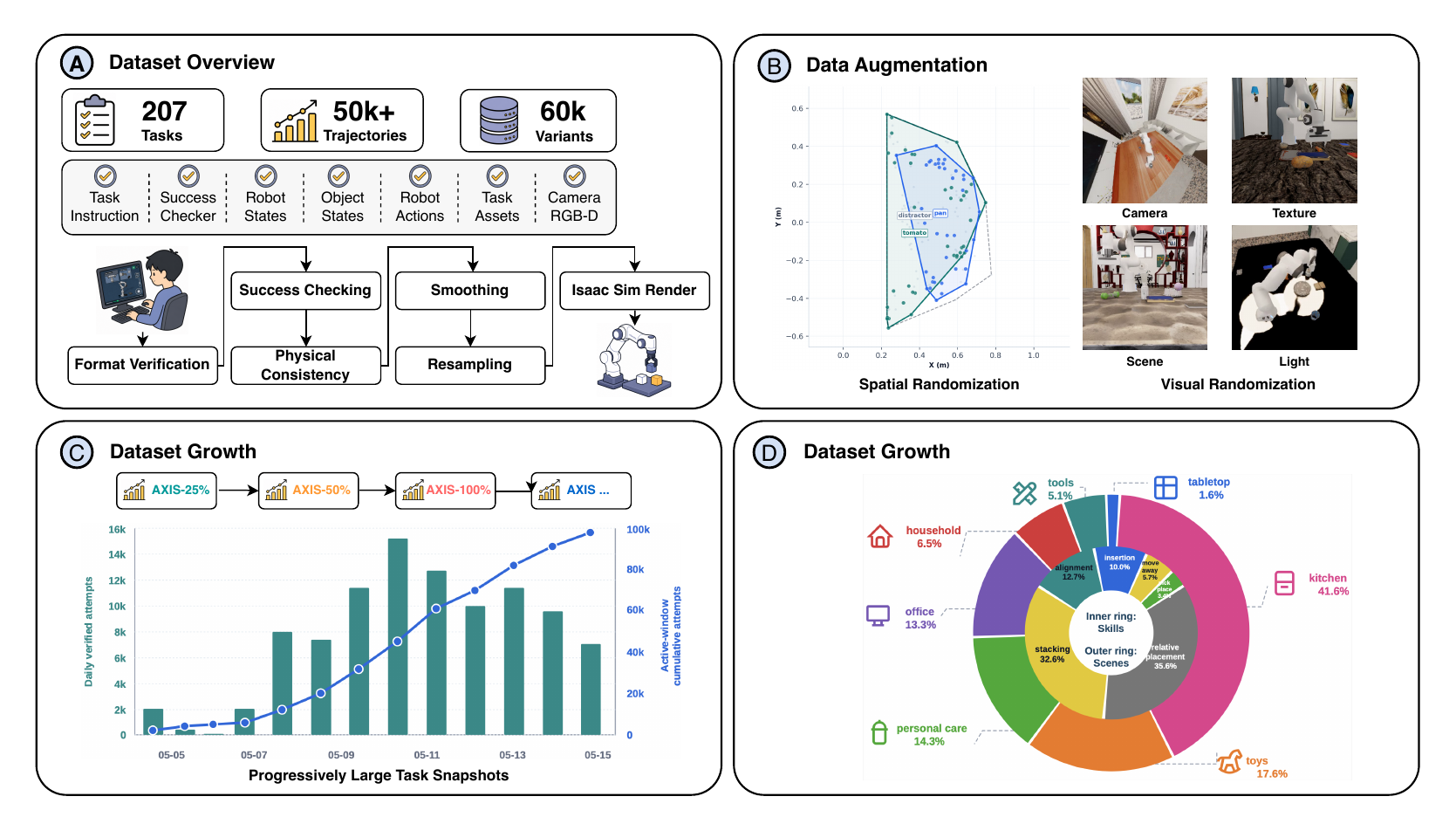}
    \vspace{-0.7cm}
    \caption{Overview of the AXIS dataset.}
    \label{fig:data_overview}
    \vspace{-0.6cm}
\end{figure}
\noindent \textbf{Dataset overview.}
AXIS is a large-scale tabletop manipulation dataset built around a Franka Research 3 robot equipped with a parallel-jaw gripper. As summarized in Fig.~\ref{fig:data_overview}, the current dataset snapshot contains 207 manipulation tasks, over 50K human demonstration trajectories, and more than 60K task or scene variants. Each task is specified by a language instruction, a parameterized simulation scene, task assets, and a structured success checker. Each demonstration trajectory contains synchronized multi-modal information, including RGB-D renderings from two camera views, robot states, object states, robot actions, task metadata, and success labels. This unified data schema enables the same trajectories to be used for imitation learning, replay-based verification, visual augmentation, and benchmark evaluation.
The task suite covers a broad range of common tabletop manipulation behaviors, including pick-and-place, relocation, stacking, sorting, pushing, pouring, insertion, alignment, and interaction with articulated or container-like objects, and the current task distribution spans seven scene categories.

\noindent \textbf{Validation and refinement.}
To ensure data quality, all demonstrations undergo a standardized validation pipeline that verifies trajectory completeness, task success, and physical consistency. Valid trajectories are subsequently smoothed and resampled to a unified control frequency, improving temporal consistency and reducing teleoperation noise. As shown in Table~\ref{tab:data_refinement}, the refinement process substantially reduces acceleration and jerk while maintaining a high replay success rate. Additional implementation details are provided in the Appendix.

\noindent \textbf{Data augmentation.}
To improve diversity and robustness, AXIS incorporates augmentation through spatial and visual randomization. These augmentations generate additional valid scene configurations while preserving task semantics and success conditions. As illustrated in Fig.~\ref{fig:data_demo}, the resulting variations cover changes in object layouts, viewpoints, textures, lighting, and backgrounds, expanding the dataset to more than 60K scene variants. This diversity makes AXIS particularly useful for studying robustness and generalization under environmental variations.

\noindent \textbf{Benchmarking platform.}
AXIS is designed as a growable benchmark for studying policy learning as data grows. The training data is organized into progressively larger task snapshots, including AXIS-25\%, AXIS-50\%, and AXIS-100\%. Each snapshot preserves the same data format, validation pipeline, task definitions, rollout budget, and success checkers, enabling controlled scaling studies in which only the amount and diversity of training data are varied.
The dataset is also continuously expandable through community-driven task and trajectory collection. In the active collection window shown in Fig.~\ref{fig:data_overview}, the number of daily verified attempts reaches approximately 15K at its peak, and the active-window cumulative number of attempts approaches 100K. These statistics indicate that AXIS is intended to support incremental dataset growth rather than a one-time static release. 

The benchmark supports two complementary evaluation modes. First, AXIS enables direct comparison across representative policy families, including conventional visuomotor imitation learning methods and vision-language-action policies. Second, AXIS supports within-embodiment scaling studies, where policies are trained on AXIS-25\%, AXIS-50\%, AXIS-100\%, and future larger snapshots to measure whether additional task coverage improves held-out manipulation generalization.
Together, these properties position AXIS not only as a large-scale manipulation dataset, but also as a long-term benchmark for studying scalable robot learning and data-driven generalization.

\vspace{-0.2cm}
\section{Experiments}
\vspace{-0.2cm}

\label{sec:experiments}

\begin{table}[t]
\centering
\scriptsize
\caption{Effect of data refinement on teleoperation trajectory quality. Relative to raw teleoperation, the final smoothed-and-resampled trajectories reduce mean acceleration and mean jerk by 63.9\% and 80.8\%, respectively.}
\label{tab:data_refinement}
\begin{tabular}{lcccc}
\toprule
Data Version & Sampling Rate & Mean Acceleration & Mean Jerk & Replay Success \\
\midrule
Raw Teleoperation & 5.0 Hz & 1.3539 & 11.5899 & 100.0\% \\
Smoothed & 5.0 Hz & 0.6382 & 2.9160 & 91.4\% \\
Smoothed + Resampled & 20 Hz & 0.4885 & 2.2243 & 86.2\% \\
\bottomrule
\end{tabular}
\end{table}

To evaluate AXIS as a growable data engine, we organize our experiments around three questions. (i)~Does continual pretraining $\pi_{0.5}$~\cite{intelligence2025pi05} on AXIS improve downstream LIBERO-Plus~\cite{libero_plus_2025} performance, and is the improvement specific to AXIS versus a matched-volume baseline drawn from RoboCasa365~\cite{robocasa365}? (ii)~Does the improvement scale with AXIS volume, and is it largest when the downstream LIBERO fine-tuning data is scarce? (iii)~On which LIBERO-Plus perturbation axes does AXIS continual pretraining help most, and does the pattern match the perturbations explicitly randomized by the AXIS IsaacSim augmentation pipeline?

\subsection{Experimental Setup}

\noindent \textbf{AXIS dataset.}
The AXIS training corpus is collected through the browser-based MuJoCo-WASM teleoperation frontend and stored in the unified AXIS trajectory format. The robot embodiment is a Franka Research 3 robot arm with a parallel-jaw gripper; tasks cover tabletop manipulation behaviors including pick-and-place, stacking and sorting, articulated object interaction, pushing, pouring, and tool-use-style manipulation. Raw demonstrations are passed through the AXIS processing pipeline (success validation, static-segment removal, Savitzky-Golay smoothing, fixed-rate resampling) and then replayed in IsaacSim under randomized visual and physical conditions. We use the AXIS-100\% snapshot as the largest continual pretraining pool, and sample uniformly without replacement at the task level to construct the 25\% and 50\% subsets used in scaling experiments.

\noindent \textbf{Training pipeline.}
All conditions share the same three-stage training pipeline: start from the released $\pi_{0.5}$ checkpoint; optionally pretrain on a sim corpus (AXIS subset or RoboCasa365 subset); fine-tune on the LIBERO trajectory set provided with LIBERO-Plus. The continual pretraining stage uses identical optimizer settings, learning rate, batch size, and gradient-step budget across all non-vanilla conditions; the fine-tuning stage uses the $\pi_{0.5}$ default LIBERO recipe across all conditions. The RoboCasa-matched control is constructed by sampling RoboCasa365 trajectories without replacement until the trajectory count equals AXIS-100\%, controlling for raw training volume.

\noindent \textbf{Evaluation protocol.}
All policies are evaluated on the LIBERO-Plus robustness suite, which extends LIBERO with systematic perturbations along seven axes (Camera, Light, Sensor Noise, Background, Layout, Language, Robot). For each (task, perturbation axis) pair we fix the rollout budget to a constant $K$ across conditions, and report mean success rate over the rollouts. As a within-distribution sanity check, we additionally report success rate on the held-out task split of AXIS-100\% itself; this confirms that AXIS continual pretraining does not destructively shift the policy away from the AXIS task distribution.

\subsection{Main Results: AXIS Continual Pretraining vs.\ Matched Sim Baseline}
\label{sec:main_result}

Our first experiment tests whether AXIS data is useful as continual pretraining for a downstream manipulation benchmark, and whether the benefit is specific to AXIS or generic to any matched sim corpus. We treat AXIS as \emph{continual pretraining} data on top of $\pi_{0.5}$~\cite{intelligence2025pi05}, fine-tune the resulting model on the standard LIBERO trajectories, and evaluate on the LIBERO-Plus robustness benchmark~\cite{libero_plus_2025}. This design holds the downstream task family, the fine-tuning data, and the evaluation distribution fixed for every method, isolating the effect of the continual pretraining corpus.

We compare four training conditions. (1)~\textbf{$\pi_{0.5}$ vanilla} performs the LIBERO fine-tuning directly from the released $\pi_{0.5}$ checkpoint and establishes the floor. (2)~\textbf{$\pi_{0.5}$ + AXIS-25\%} pretrains on a 25\% random subset of AXIS-100\% before LIBERO fine-tuning, providing a low-volume scaling point. (3)~\textbf{$\pi_{0.5}$ + AXIS-100\%} pretrains on the full AXIS-100\% snapshot and is our headline condition. (4)~\textbf{$\pi_{0.5}$ + RoboCasa-matched} pretrains on a RoboCasa365~\cite{robocasa365} subset matched in trajectory count to AXIS-100\%, controlling for the possibility that any comparably-sized Franka-sim training corpus would yield similar gains. Together, conditions (1)-(3) measure whether AXIS helps and whether more AXIS helps more; condition (4) measures whether the gain is AXIS-specific or generic to simulation continual pretraining at this volume.
All conditions share the same downstream LIBERO fine-tuning recipe ($\pi_{0.5}$ default optimizer, batch size, schedule, and step budget), the same LIBERO trajectory set, the same LIBERO-Plus evaluation protocol, and the same number of rollouts per task and perturbation condition. Training hyperparameters are also matched across conditions (2)-(4): identical learning rate, batch size, gradient-step budget, and optimizer state initialization.

\begin{table}[!tbp]
    \centering
    \scriptsize
    \caption{Main results. All conditions start from the released $\pi_{0.5}$ checkpoint, optionally pretrain on the listed corpus, then fine-tune on LIBERO trajectories using identical hyperparameters, and evaluate on LIBERO-Plus. The RoboCasa-matched control uses the same trajectory count as AXIS-100\%. Per-perturbation columns report success rate on the corresponding LIBERO-Plus axis.}
    \label{tab:axis_pretrain_main}
    \setlength{\tabcolsep}{3pt}
    \renewcommand{\arraystretch}{1.10}
    \begin{tabular*}{\linewidth}{@{\extracolsep{\fill}}lccccccccc}
        \toprule
        Model & \# pretrain demos & Overall $\uparrow$ & Cam. & Light & Noise & B.G. & Layout & Lang. & Robot \\
        \midrule
        $\pi_{0.5}$ vanilla & 0 & 83.9 & 72.5 & 98.2 & 82.5 & 94.4 & 82.9 & 89.6 & 74.4 \\
        $\pi_{0.5}$ + AXIS-25\% & $0.25\,N_{\mathrm{AXIS}}$ & 84.7 & 77.5 & 91.2 & 86.2 & 100.0 & 85.5 & 81.8 & 76.9 \\
        $\pi_{0.5}$ + AXIS-50\% & $0.50\,N_{\mathrm{AXIS}}$ & 85.7 & 68.8 & 98.2 & 91.2 & 96.3 & 88.2 & 84.4 & 79.5 \\
        $\pi_{0.5}$ + AXIS-100\% (ours) & $N_{\mathrm{AXIS}}$ & 88.8 & 83.8 & 96.5 & 96.2 & 98.1 & 85.5 & 88.3 & 78.2 \\
        $\pi_{0.5}$ + RoboCasa-matched (ctrl.) & $N_{\mathrm{AXIS}}$ & 57.5 & 35.2 & 79.5 & 63.2 & 81.7 & 68.0 & 49.2 & 39.4 \\
        \bottomrule
    \end{tabular*}
\end{table}

The key question addressed by Table~\ref{tab:axis_pretrain_main} is whether AXIS continual pretraining provides benefits beyond the existing continual pretraining of $\pi_{0.5}$, and whether those benefits are specific to AXIS data. AXIS continual pretraining consistently improves downstream performance over the vanilla $\pi_{0.5}$ baseline, with gains increasing as more AXIS data is added. The performance gap between AXIS-25\% and AXIS-100\% further suggests that these benefits have not yet saturated, indicating that additional AXIS data continues to improve robustness on downstream manipulation tasks. 
The matched-volume RoboCasa365 control tests whether these gains arise from AXIS itself or simply from additional simulation data. Despite using the same training volume, RoboCasa-matched underperforms both AXIS variants across nearly all perturbation categories and on the overall benchmark. This suggests that the gains are not explained by dataset size alone, but by properties unique to the AXIS pipeline, including crowdsourced task creation, diverse demonstrations, and large-scale environment randomization.

Overall, the results support AXIS as a scalable data engine: increasing AXIS data yields consistent downstream improvements, and those gains stem from the diversity and coverage of the AXIS pipeline rather than training volume alone.

\subsection{Per-Perturbation Robustness Analysis}

To understand where the AXIS dataset helps and where it does not, we report the per-perturbation breakdown of the AXIS-100\% condition against vanilla $\pi_{0.5}$ and the RoboCasa-matched control at the full LIBERO fine-tune budget. The LIBERO-Plus perturbation axes probe different distribution shifts: \emph{Camera} varies viewpoint, \emph{Light} varies illumination, \emph{Sensor Noise} adds photometric perturbations (Gaussian noise and contrast jitter) to the camera observations, \emph{Background} varies scene appearance and table/wall textures, \emph{Layout} varies object initial positions and orientations, \emph{Language} varies task wording, and \emph{Robot} varies the robot initial pose. Different training corpora are expected to help unequally across these axes depending on what their training distribution covers; AXIS's IsaacSim augmentation specifically targets camera, light, texture, and layout perturbations, so we expect the strongest relative gains on those axes.

\begin{table}[!tbp]
    \centering
    \scriptsize
    \caption{Per-perturbation LIBERO-Plus breakdown at the full LIBERO fine-tune budget. Each row reports success rate (\%) on the corresponding perturbation axis. $\Delta_{\mathrm{van}}$ is the absolute improvement of AXIS-100\% over $\pi_{0.5}$ vanilla. $\mathrm{AXIS}{-}\mathrm{RC}$ is the difference between AXIS-100\% and the matched-volume RoboCasa365 control, measuring the gain over a same-size sim corpus rather than per-trajectory superiority.}
    \label{tab:task_family_breakdown}
    \setlength{\tabcolsep}{3pt}
    \renewcommand{\arraystretch}{1.10}
    \begin{tabular*}{\linewidth}{@{\extracolsep{\fill}}lcccccc}
        \toprule
        Perturbation axis & $\pi_{0.5}$ vanilla & $+$ AXIS-25\% & $+$ AXIS-100\% & $+$ RoboCasa-m. & $\Delta_{\mathrm{van}}$ & $\mathrm{AXIS}{-}\mathrm{RC}$ \\
        \midrule
        Camera & 72.5 & 77.5 & 83.8 & 35.2 & +11.3 & +48.6 \\
        Light & 98.2 & 91.2 & 96.5 & 79.5 & $-1.7$ & +17.0 \\
        Sensor Noise & 82.5 & 86.2 & 96.2 & 63.2 & +13.7 & +33.0 \\
        Background & 94.4 & 100.0 & 98.1 & 81.7 & +3.7 & +16.4 \\
        Layout & 82.9 & 85.5 & 85.5 & 68.0 & +2.6 & +17.5 \\
        Language & 89.6 & 81.8 & 88.3 & 49.2 & $-1.3$ & +39.1 \\
        Robot & 74.4 & 76.9 & 78.2 & 39.4 & +3.8 & +38.8 \\
        \midrule
        \textbf{Overall} & 83.9 & 84.7 & 88.8 & 57.5 & +4.9 & +31.3 \\
        \bottomrule
    \end{tabular*}
    \vspace{-1.5em}
\end{table}

We use Table~\ref{tab:task_family_breakdown} to understand where AXIS dataset helps and where it does not. As expected from the AXIS augmentation design, the largest gains occur on visual and scene-related perturbations, including camera viewpoint, sensor noise, background appearance, and object layout, showing that IsaacSim randomization transfers effectively to corresponding robustness challenges in LIBERO-Plus.
In addition, AXIS also improves performance on language and robot-pose variation, despite not explicitly augmenting these factors. This suggests that AXIS provides more than perturbation-specific robustness: exposure to diverse tasks, behaviors, and interaction contexts improves the underlying manipulation representations, leading to broader gains under distribution shift.
Comparisons with the RoboCasa365 baseline of the same volume reinforce this conclusion. 
AXIS outperforms the RoboCasa365 control across nearly all perturbation axes, indicating that its benefits stem not only from additional simulation data but also from the diversity of environments, tasks, and demonstrations it provides. Overall, AXIS improves robustness broadly, with particularly strong gains on the visual and geometric shifts it was designed to address.

\section{Discussion}

\noindent \textbf{Why growable datasets matter.}
Static datasets are valuable for benchmarking, but robot manipulation learning also benefits from data sources that can grow with model capability and observed failures. AXIS turns data collection into an iterative feedback loop: our task generation  module expands semantic diversity through new manipulation problems, community teleoperation adds behavioral diversity through varied human strategies and corrections, and visual and physics augmentation expands environmental diversity through changes in appearance, scene configuration, and dynamics. In this sense, dataset growth becomes a mechanism for improving policy robustness rather than a one-time release process.

\noindent \textbf{Community collection and data quality.}
Community-driven collection improves scale and accessibility, but introduces significant variation in operator skill, motion smoothness, strategy, and task completion quality. Useful variation exposes policies to diverse grasps, recoveries, and correction patterns, while harmful variation appears as corrupted states, idle segments, failed rollouts, discontinuities, or unstable actions. AXIS addresses this tradeoff by making validation, cleaning, smoothing, replay, and augmentation core stages of dataset construction, unifying broad participation with the quality control needed for downstream policy learning.

\section{Conclusion}
We introduced AXIS, a growable community-driven data engine and benchmark for scalable learning for robot manipulation. AXIS combines browser-based MuJoCo-WASM teleoperation, automated task generation, distributed demonstration collection, standardized trajectory processing, IsaacSim-based augmentation, and fixed-protocol evaluation. It constructs a training-ready dataset that can expand through new tasks, demonstrations, and augmented conditions, while enabling controlled comparisons across vision-language-action policies. More broadly, AXIS points toward continuously extensible robot data pipelines that connect task generation, community collection, policy training, and failure-driven improvement.

\noindent \textbf{Limitations and Future Work.} AXIS currently focuses on simulated Franka tabletop manipulation, and sim-to-real transfer remains an open challenge. Future work will expand AXIS to include more robot embodiments, richer sensing modalities, longer-horizon tasks, finer-grained annotations, and active failure-driven data collection to support more generalizable and scalable robot learning.

\newpage
\section*{Author Contributions \& Acknowledgement}

\noindent \textbf{Core Contributors:} Mengfei Zhao, Dihong Huang, Yikai Tang, Peihao Li, Mingxuan Yan, Ruiqi Zhuang, Yanjia Huang, Jie Wang, Hai Zhai, Tony Zhou, Rui Zhang, Zhexi Luo, and Yuchen Huang. These project members made substantive contributions to the scientific outcomes, including data collection, system development, experiments, result analysis, and paper writing.

\noindent \textbf{Principal Investigators (PIs):} Jianfei Yang and Jiachen Li. The PIs provided project supervision and research direction.

\noindent \textbf{Data Collection Contributors:} We thank over 70,000 members of the Axis Robotics community who contributed demonstrations through the AXIS web-based teleoperation platform. Their participation is what makes AXIS a continuously growing, community-driven data engine.

\noindent \textbf{Industry Support:} We thank the collaborators at Axis Robotics for data contribution, engineering support, infrastructure assistance, and collaboration on dataset development.


\bibliography{example}
\clearpage

\appendix
\makeatletter
\newcommand{\apxtableofcontents}{%
  \section*{Appendix Contents}%
  \@starttoc{apx}%
  \vspace{0.5em}%
}

\newcommand{\apxaddsection}[1]{%
  \addcontentsline{apx}{section}{\protect\numberline{\thesection}#1}%
}
\newcommand{\apxaddsubsection}[1]{%
  \addcontentsline{apx}{subsection}{\protect\numberline{\thesubsection}#1}%
}

\let\apx@old@section\section
\renewcommand{\section}{\@ifstar{\apx@old@section*}{\apx@section}}
\newcommand{\apx@section}[1]{\apx@old@section{#1}\apxaddsection{#1}}

\let\apx@old@subsection\subsection
\renewcommand{\subsection}{\@ifstar{\apx@old@subsection*}{\apx@subsection}}
\newcommand{\apx@subsection}[1]{\apx@old@subsection{#1}\apxaddsubsection{#1}}
\makeatother

\apxtableofcontents

\raggedbottom
\setlength{\textfloatsep}{8pt}
\setlength{\intextsep}{8pt}
\setlength{\floatsep}{8pt}

\newpage

\section{Dataset Comparison}
\label{sup:dataset-comparison}

\begin{table}[H]
\centering
\small
\caption{Comparison of representative robot manipulation datasets in terms of collection setting, scale, task coverage, and supervision.}
\label{tab:dataset_comparison}
\setlength{\tabcolsep}{4pt}
\renewcommand{\arraystretch}{1.12}

\begin{tabularx}{\textwidth}{
@{}
>{\raggedright\arraybackslash}p{0.16\textwidth}
>{\centering\arraybackslash}p{0.05\textwidth}
>{\raggedright\arraybackslash}p{0.12\textwidth}
>{\raggedright\arraybackslash}p{0.16\textwidth}
>{\raggedright\arraybackslash}p{0.16\textwidth}
>{\raggedright\arraybackslash}X
@{}}
\toprule
\textbf{Dataset} &
\textbf{Year} &
\textbf{Setting} &
\textbf{Scale} &
\textbf{Task Coverage} &
\textbf{Data / Supervision} \\
\midrule

RoboNet~\cite{dasari2020robonetlargescalemultirobotlearning} &
2019 &
Real robot &
162K trajectories &
Diverse manipulation tasks &
RGB videos, robot actions, gripper states, and action trajectories \\

LIBERO~\cite{NEURIPS2023_8c3c6668} &
2023 &
Simulation &
6.5K trajectories &
130 tasks &
RGB video, proprioception, language instructions, and expert demonstrations \\

BridgeData V2~\cite{pmlr-v229-walke23a} &
2023 &
Real robot &
60K trajectories; 24 environments &
13 skills &
RGB videos, proprioception, goal images, and language instructions \\

Open X-Embodiment~\cite{10611477} &
2024 &
Real robot &
1M+ episodes; 60 datasets &
500+ skills &
RGB videos, end-effector poses, language instructions, and action trajectories \\

DROID~\cite{khazatsky2025droidlargescaleinthewildrobot} &
2024 &
Real robot &
76K trajectories; 564 scenes &
86 tasks &
RGB-D videos, proprioception, teleoperated actions, and language instructions \\

RoboMIND~\cite{wu2025robomind} &
2024 &
Real robot &
107K trajectories; 96 objects &
279 tasks &
RGB-D videos, proprioception, language, teleoperation, and failure demonstrations \\

PartInstruct~\cite{yin2025partinstruct} &
2025 &
Simulation &
513 objects; 1,302 tasks; 10K+ demos &
16 part-level task classes &
Language, part-level 3D object data, skill sequences, and expert demonstrations \\

\midrule
\textbf{Ours} &
\textbf{2026} &
\textbf{Web-based simulation} &
\textbf{50,129 episodes} &
\textbf{207 tasks} &
\textbf{Browser-based episodes with scalable user-contributed manipulation data} \\

\bottomrule
\end{tabularx}
\end{table}

Table~\ref{tab:dataset_comparison} summarizes representative robot manipulation datasets spanning real-world robotics, simulation environments, and large-scale embodied learning benchmarks. Existing datasets generally emphasize either large-scale real-world data collection (e.g., RoboNet, Open X-Embodiment, DROID, and RoboMIND) or highly controlled simulation benchmarks with expert demonstrations (e.g., LIBERO and PartInstruct). While these datasets have substantially advanced robot learning, their collection pipelines often require specialized robotic hardware, expert operators, or centralized infrastructure, limiting scalability and accessibility.

In contrast, our dataset adopts a web-based simulation paradigm that enables robot manipulation data collection directly through standard web browsers. The framework emphasizes three key properties: scalability, continual growth, and community participation. The absence of physical robots, specialized infrastructure, and expert operators significantly lowers participation barriers and enables large-scale data contribution from a broad user community. New tasks, objects, and interaction scenarios can be incorporated continuously as the community expands. The current release contains 50,129 episodes spanning 207 manipulation tasks, illustrating the potential of community-driven data generation for robotic manipulation. We envision this framework as a complementary direction to existing real-world and simulator-based datasets and as a sustainable pathway toward ever-growing robot manipulation corpora.

\section{Web-Infra: Browser-Based Infrastructure}
\label{sup:web-based-infrastructure}

AXIS is a cross-simulator, cross-machine infrastructure for scalable robot demonstration collection, trajectory processing, and policy evaluation. The key system decision is to separate interactive data collection from compute-heavy backend processing. Contributors teleoperate robots in a browser-based MuJoCo WebAssembly frontend using commodity devices, while uploaded demonstrations are routed to backend machines for validation, cleaning, replay, rendering, augmentation, and export to downstream policy-learning pipelines. This split lowers the entry cost for data contributors without forcing the full pipeline to inherit the limits of a lightweight browser simulator.

At the pipeline level, AXIS follows six stages: browser-based teleoperation and logging, trajectory upload in a unified format, backend replay and realistic augmentation, data cleaning and curation, model training with sim-to-real evaluation, and real-robot deployment. The pipeline is intentionally asymmetric: latency-sensitive interaction remains on commodity web clients, while throughput-sensitive rendering, replay, model training, and evaluation run on backend machines. Data rendering is performed on 8$\times$RTX 4090 GPUs, while all model training and evaluation are performed on 8$\times$A100 GPUs. This organization lets the same demonstration serve multiple purposes, including task verification, cleaned control supervision, realistic RGB rendering, domain-randomized augmentation, downstream policy learning, and evaluation.

The infrastructure is organized around shared task and trajectory abstractions. At the task level, environments expose common observation, action, reset, step, and success-checking interfaces, so a task collected in MuJoCo-WASM can later be verified, replayed, or rendered in a backend simulator without rewriting the task logic. At the trajectory level, each demonstration stores metadata, time-series signals, and episode-level annotations in a unified format. Metadata include the task name, environment identifier, robot embodiment, simulator version, contributor identifier when available, and timestamps. Time-series fields include actions, robot states, observations, and optional object states or visual observations. Episode-level fields include success labels, segmentation markers, failure modes when available, and links to the task-specific success checker.

\begin{table}[H]
\centering
\small
\caption{Functional decomposition of the Web-Infra pipeline.}
\label{tab:webinfra_pipeline}
\setlength{\tabcolsep}{4pt}
\renewcommand{\arraystretch}{1.08}
\begin{tabularx}{\textwidth}{@{}p{0.22\textwidth}p{0.20\textwidth}X@{}}
\toprule
\textbf{Stage} & \textbf{Execution site} & \textbf{Role in the pipeline} \\
\midrule
Task deployment & Browser client & Loads a parameterized task, robot model, scene objects, and success checker into the MuJoCo-WASM frontend. \\
Teleoperation & Browser client & Maps keyboard, mouse, joystick, or gamepad commands to end-effector targets and joint-level robot actions. \\
Logging & Browser client & Buffers time-aligned state-action samples inside the physics loop and serializes successful attempts with task metadata. \\
Upload and indexing & Backend service & Stores trajectories with versioned metadata so they can be retrieved for validation, replay, rendering, and training. \\
Cleaning and replay & Backend machines & Filters invalid episodes, smooths and resamples motion, verifies success, and reconstructs trajectories for rendering. \\
Learning export & Backend machines & Packages cleaned and rendered data for imitation learning and VLA training. \\
\bottomrule
\end{tabularx}
\end{table}

The browser frontend is implemented as an end-to-end teleoperation interface that covers task selection, interactive 3D visualization, camera control, control-state display, and automatic trajectory packaging. The core dashboard acts as a heads-up display over the MuJoCo-WASM and WebGL viewport. A top-level status area reports real-time metrics such as elapsed time and rendering rate, while the control panel groups gripper motion, Cartesian movement, rotation, action buttons, and camera controls so new contributors can quickly understand the available inputs. The interface supports keyboard, mouse, virtual joystick, and gamepad control, making data collection compatible with heterogeneous personal computers rather than specialized lab hardware.

To keep browser control responsive, AXIS uses an orchestrated runtime rather than a monolithic UI loop. MuJoCo physics stepping and Three.js rendering are separated from the main React UI thread. A lightweight broadcaster sends occasional status updates to the interface, while the trajectory manager buffers state-action samples directly inside the high-frequency control loop. This design avoids frequent UI re-rendering during teleoperation and keeps the logged samples aligned with the simulator state. Once the embedded checker verifies task completion, the buffered samples and task metadata are packaged into a standardized trajectory record for backend upload.

Operator commands are mapped to six-degree-of-freedom Cartesian end-effector targets and converted to joint-level robot commands through inverse kinematics. For more complex end-effectors, AXIS also supports interpolation-based grasp control. Let $\mathbf{q}_{\mathrm{rest}} \in \mathbb{R}^{n}$ denote the fully open hand pose and $\mathbf{q}_{\mathrm{tmpl}} \in \mathbb{R}^{n}$ denote a grasp template, such as a power, precision, or functional grasp. The commanded hand pose is
\begin{equation}
  \mathbf{q}(t) =
  \mathbf{q}_{\mathrm{rest}} +
  \alpha(t)\left(\mathbf{q}_{\mathrm{tmpl}}-\mathbf{q}_{\mathrm{rest}}\right),
\end{equation}
where $\alpha(t) \in [0, 1]$ is controlled by the operator. This collapses a high-dimensional hand configuration, often with more than 16 actuated joints, into a single scalar control axis. New grippers or dexterous hands can therefore be integrated by defining rest poses and task-specific template libraries, without changing the core teleoperation controller.

\section{TaskGen: Task Generation and Deployment}
\label{sup:taskgen}

TaskGen converts a high-level manipulation instruction into a simulation-ready task instance. Given an instruction such as \textit{pick up the toy car on the table}, a task manager decomposes the request into task, scene, and object configurations. The task configuration specifies the manipulation goal and required object interactions. The scene configuration defines workspace context such as supporting surfaces, clutter level, and spatial constraints. The object configuration determines which object models appear in the scene, their semantic roles, and whether they are required for task completion or included as decorative context. Each task is also assigned a difficulty level from 1 to 5, which controls the number of objects, the tightness of spatial constraints, and the overall scene complexity.

\begin{figure}[t]
    \centering
    \includegraphics[width=\linewidth]{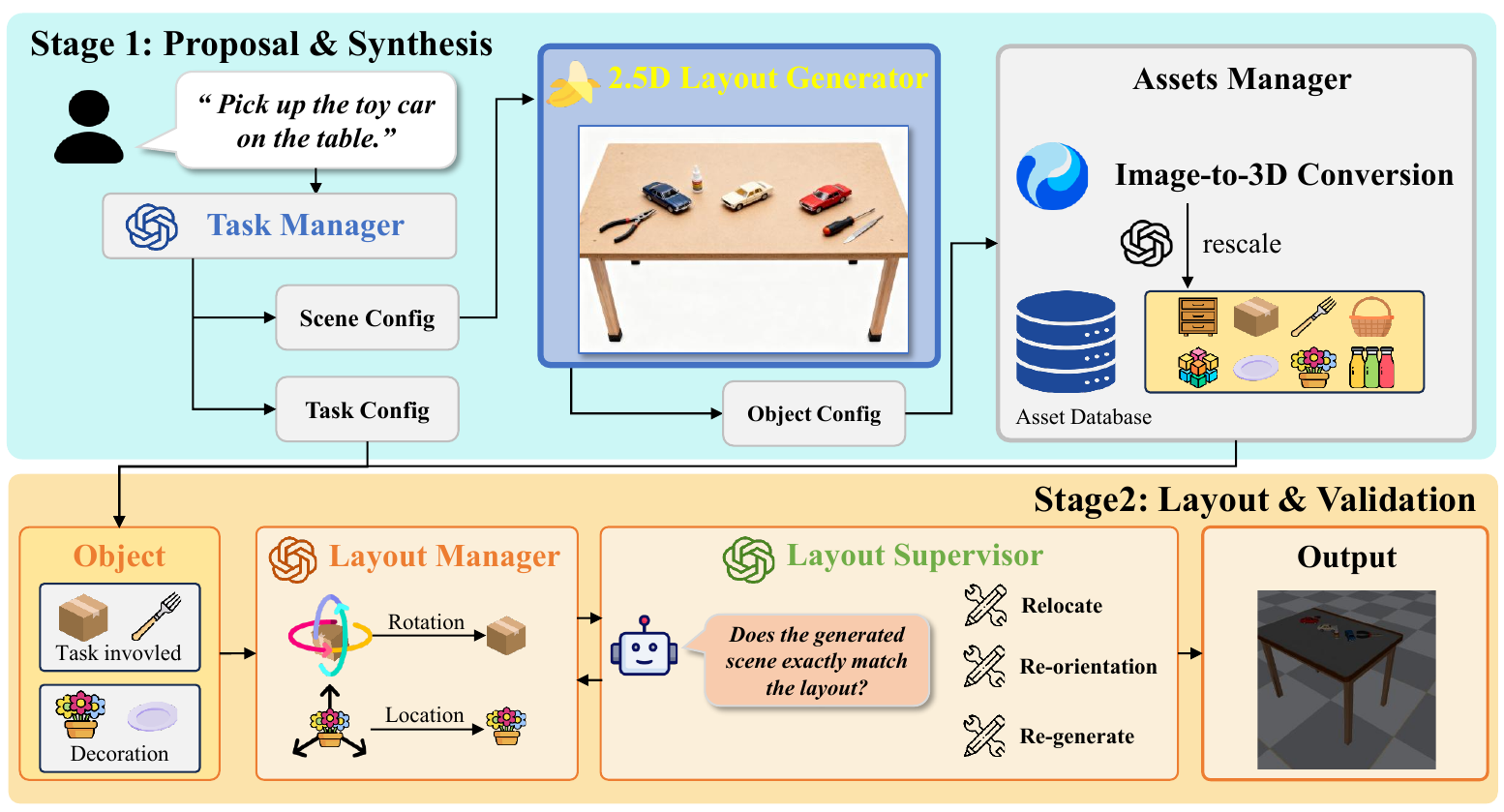}
    \caption{TaskGen pipeline overview. A language instruction is converted into task, scene, and object specifications; object models are retrieved or generated and normalized; a layout is proposed, instantiated, checked, and deployed.}
    \label{fig:sup_taskgen_pipeline}
\end{figure}

TaskGen uses an object-model manager to populate scenes with diverse objects. Object models can be retrieved from an existing database or generated through an image-to-3D conversion pipeline. Because generated meshes may have inconsistent physical scales, the object-model manager normalizes them to plausible dimensions before adding them to the task database. This normalization step is important for downstream teleoperation and replay: object size affects reachability, grasp feasibility, collision behavior, and the validity of geometric success conditions.

Conditioned on the task specification, selected object models, and difficulty level, a 2.5D layout generator proposes an initial scene arrangement. The generator describes which objects should appear on the workspace and their approximate spatial relationships. A layout manager then instantiates the proposal into a full 3D environment by assigning each object a precise location and orientation. A layout supervisor verifies whether the instantiated scene satisfies the intended task constraints. When the supervisor detects an inconsistency, such as an unreachable object, an invalid support relation, or a mismatch between the language goal and object placement, the system iteratively corrects the scene through relocation, re-orientation, or partial regeneration.

\begin{table}[H]
\centering
\small
\caption{TaskGen pipeline components and their outputs.}
\label{tab:taskgen_pipeline}
\setlength{\tabcolsep}{4pt}
\renewcommand{\arraystretch}{1.08}
\begin{tabularx}{\textwidth}{@{}p{0.20\textwidth}p{0.26\textwidth}X@{}}
\toprule
\textbf{Component} & \textbf{Input} & \textbf{Output / validation role} \\
\midrule
Task manager & Language instruction, difficulty level & Structured task, scene, and object configurations with semantic object roles. \\
Object-model manager & Object requirements and model database & Retrieved or image-to-3D generated object models normalized to plausible physical scale. \\
2.5D layout generator & Task specification, object models, clutter settings & Initial object arrangement over the workspace, including task-involved and decorative objects. \\
Layout manager & 2.5D proposal and object geometry & Full 3D scene with object poses, orientations, support relations, and workspace placement. \\
Layout supervisor & Instantiated scene and task constraints & Validated scene or corrective actions such as relocation, re-orientation, or partial regeneration. \\
Success checker & Task goal and object interactions & Automatic completion test reused during browser collection and backend verification. \\
\bottomrule
\end{tabularx}
\end{table}

After a valid layout is produced, TaskGen defines distributions over initial conditions to support scenario sampling. These distributions vary object poses, clutter configurations, and camera viewpoints while preserving the core task semantics. Each generated instance is paired with a task-specific success checker that encodes completion using geometric constraints and object interactions. The checker is used both during browser teleoperation, where it terminates and packages successful demonstrations, and during backend validation, where it helps verify whether uploaded trajectories remain task-consistent after transfer.

The output of TaskGen has three parts: a parameterized task definition describing scene configuration and object layout, a unified observation and action interface for execution across simulators, and a task-specific success checker for automatic evaluation. This decomposition makes task growth continuous rather than one-off. New language instructions, objects, layouts, and difficulty settings can be added to the task pool, deployed to the browser, collected by contributors, and then replayed or rendered through the same backend pipeline.

\section{Data Cleaning and Trajectory Refinement}
\label{sup:data-cleaning}

Raw web demonstrations require filtering and refinement before they can be used for policy learning. Browser teleoperation can introduce jitter, stalled samples, unintended oscillations, low-frequency sampling artifacts, corrupted records, and occasional task-completion mismatches. AXIS therefore treats data cleaning as a production pipeline with two goals: filter unusable episodes and refine borderline episodes into temporally smooth, replayable trajectories.

\begin{algorithm}[t]
\caption{Trajectory cleaning and refinement.}
\label{alg:sup-cleaning}
\begin{algorithmic}[1]
\Require Raw episode $S$, static threshold $\epsilon$, raw timestep $\Delta t_{\mathrm{raw}}$, target timestep $\Delta t$, Savitzky--Golay parameters $W, P$
\Ensure Smoothed and resampled episode $S_{\mathrm{final}}$

\State \textbf{\textit{Step 1: Validation and segmentation}}
\State Reject $S$ if required fields, finite values, length consistency, physical continuity, or task success checks fail.
\State $S_{\mathrm{clean}} \gets \Call{RemoveStaticSegments}{S, \epsilon}$
\State $T_{\mathrm{clean}} \gets \Call{RebuildUniformTime}{S_{\mathrm{clean}}, \Delta t_{\mathrm{raw}}}$
\State $Q_{\mathrm{clean}} \gets \Call{RobotJointSequence}{S_{\mathrm{clean}}}$

\State \textbf{\textit{Step 2: Robot-motion smoothing}}
\State $Q_{\mathrm{smooth}} \gets \Call{SavitzkyGolay}{Q_{\mathrm{clean}}, W, P}$
\Statex \Comment{Discrete gripper transitions can be excluded from continuous smoothing when needed.}

\State \textbf{\textit{Step 3: Fixed-frequency resampling}}
\State $T_{\mathrm{new}} \gets \Call{Arange}{T_{\mathrm{clean}}[0], T_{\mathrm{clean}}[-1], \Delta t}$
\State $Q_{\mathrm{final}} \gets \Call{CubicSpline}{T_{\mathrm{clean}}, Q_{\mathrm{smooth}}}(T_{\mathrm{new}})$

\State \textbf{\textit{Step 4: Extra-state alignment and export}}
\State $S_{\mathrm{final}} \gets \Call{AlignNonRobotState}{T_{\mathrm{new}}, Q_{\mathrm{final}}, S_{\mathrm{clean}}}$
\Return $S_{\mathrm{final}}$
\end{algorithmic}
\end{algorithm}

The validation stage removes trajectories that violate basic structural, numerical, physical, or task-level checks. Structural checks reject demonstrations with missing required fields, incomplete metadata, corrupted records, or length mismatches between state and action streams. Numerical checks reject non-finite values and anomalous records. Physical checks compute successive changes in robot states and reject trajectories whose deltas exceed predefined plausible thresholds, which catches extreme discontinuities and invalid state transitions. Task-level checks remove failed episodes and can re-verify nominal success labels using the task-specific checker in the backend simulation environment.

After validation, the pipeline removes uninformative static segments. Samples are classified as static when the absolute variation across all robot joints is below a configured threshold; in the current processing pipeline, this threshold is $5 \times 10^{-3}$. Removing these samples eliminates artificial pauses and idle segments caused by human hesitation, while preserving the portions of the episode that contain meaningful control changes. The remaining trajectory is then re-timed to form a consistent temporal sequence for smoothing and resampling.

Continuous robot action or joint trajectories are smoothed with a Savitzky-Golay filter. The current implementation uses a window size of 15 and a polynomial order of 3 for robot motion smoothing, while discrete gripper transitions can be excluded from the smoothing pass so that open-close events do not become physically ambiguous. This stage reduces high-frequency teleoperation artifacts such as abrupt corrections and small oscillations without replacing the original demonstration with an unrelated planned path.

Because browser demonstrations may be recorded at a low or non-uniform rate, cleaned trajectories are resampled to a fixed control frequency with cubic spline interpolation. The web interface can produce trajectories at only 6--8 Hz, while downstream control and policy training require a higher frequency; the current target frequency is 20 Hz. Robot states and continuous action targets are interpolated after smoothing. Non-robot states, such as object poses, bypass the action-smoothing filter and are temporally aligned through interpolation only. The resulting trajectory remains faithful to the demonstrated task execution while becoming smoother, denser, and better aligned with replay and policy-learning requirements.

\begin{table}[t]
\centering
\caption{Quantitative evaluation of trajectory optimization across 16 LIBERO tasks. Smoothness (Max Acceleration and Max Jerk) is compared before and after the proposed processing.}
\label{tab:libero_metrics_comparison}
\footnotesize
\setlength{\tabcolsep}{3pt}
\renewcommand{\arraystretch}{1.1}
\begin{tabularx}{\linewidth}{@{}>{\raggedright\arraybackslash}X c c c c c c@{}}
\toprule
\multirow{3}{*}{\textbf{LIBERO Task}} & \multicolumn{4}{c}{\textbf{Smoothness}} & \multirow{3}{*}{\textbf{Pos. Dev. (m)}} & \multirow{3}{*}{\textbf{Removed Ratio}} \\
\cmidrule(lr){2-5}
& \multicolumn{2}{c}{\textbf{Mean Acc.}} & \multicolumn{2}{c}{\textbf{Mean Jerk}} & & \\
\cmidrule(lr){2-3} \cmidrule(lr){4-5}
& Before & After & Before & After & & \\
\midrule
Task 1: Place Black Bowl on Top of cabinet  & 0.2458 & 0.0738 & 2.6425 & 1.3363 & 0.0235 & 6.46\% \\
Task 2: Place Rear Butter in Cabinet Top Drawer and Close It  & 0.3067 & 0.0943 & 3.7395 & 1.6208 & 0.0654 & 5.73\% \\
Task 3: Place the black bowl on the plate  & 0.3559 & 0.1011 & 5.1358 & 2.3634 & 0.0201 & 7.17\% \\
Task 4: Place the black bowl on top of the cabinet  & 0.3225 & 0.0859 & 4.6724 & 2.0785 & 0.0142 & 3.77\% \\
Task 5: Place the frying pan on the stove  & 0.1832 & 0.1055 & 2.4771 & 1.1374 & 0.0106 & 1.24\% \\
Task 6: Place the moka pot on the stove  & 0.1819 & 0.1090 & 2.5239 & 1.6784 & 0.0064 & 1.94\% \\
Task 7: Turn on the stove  & 0.2689 & 0.1295 & 3.8291 & 2.3017 & 0.0142 & 3.81\% \\
Task 8: Close Cabinet Bottom Drawer  & 0.1574 & 0.0741 & 1.9118 & 0.6659 & 0.0057 & 4.59\% \\
Task 9: Place the black bowl into the cabinet's bottom drawer  & 0.1551 & 0.0913 & 1.8665 & 1.3193 & 0.0161 & 2.03\% \\
Task 10: Place Wine Bottle on Wine Rack & 0.1841 & 0.1114 & 2.0298 & 1.0022 & 0.0625 & 4.86\% \\
Task 11: Close Cabinet Top Drawer & 0.1224 & 0.0683 & 1.5424 & 0.5822 & 0.0053 & 2.62\% \\
Task 12: Place the black bowl into the cabinet's top drawer & 0.2109 & 0.1418 & 2.6997 & 2.1633 & 0.0961 & 11.4\% \\
Task 13: Place Black Bowl on Plate & 0.1682 & 0.0816 & 2.2262 & 1.1552 & 0.2511 & 2.14\% \\
Task 14: Place the black bowl on top of the cabinet & 0.1375 & 0.0736 & 1.8499 & 0.9924 & 0.0108 & 6.73\% \\
Task 15: Place the right moka pot on the stove & 0.1890 & 0.1130 & 2.5768 & 1.4929 & 0.0134 & 3.19\% \\
Task 16: Turn off the stove & 0.2149 & 0.1213 & 2.8268 & 1.5823 & 0.0219 & 8.93\% \\
\midrule
\textbf{Average}       & \textbf{0.2128} & \textbf{0.0985} & \textbf{2.7844} & \textbf{1.4670} & \textbf{0.0398} & \textbf{4.79\%} \\
\bottomrule
\end{tabularx}
\end{table}

The cleaned output serves as the handoff point to later rendering and augmentation stages. A retained episode contains validated metadata, temporally aligned robot and object states, smoothed robot motion, success annotations, and episode boundaries. The same cleaned record can be replayed for verification, rendered into RGB observation streams, perturbed through IsaacSim scene or camera randomization, or exported for imitation-learning and VLA training. In this way, data cleaning is not only a quality-control step but also the interface that makes web-collected demonstrations usable by the larger AXIS dataset and model-training pipeline.

\section{AXIS Data Rendering}
\label{sup:axis-rendering}

AXIS converts verified task trajectories into LIBERO-style visual imitation-learning data through state replay. The rendering source is not an action log that must be re-executed through a simulator. Instead, each eligible attempt is represented as packed simulator state, which provides the robot and object states required for timestep-by-timestep replay. This distinction matters because an action rollout can diverge from the originally verified trajectory under small simulator, contact, or controller differences, while state replay keeps the rendered visual stream aligned with the successful task execution.

Task admission is conservative. A task must have enough verified attempts both remotely and locally, and each selected attempt must be long enough to support action-chunk training. Candidate attempts are ranked by shortest verified simulation time and then by attempt id, which biases the rendered corpus toward compact successful demonstrations. The default setting renders all eligible attempts for an admitted task, so the data-generation stage emphasizes coverage once the verification threshold has been met.

Every rendered episode contains two RGB observation streams: a third-view camera that observes the workspace from outside the robot and a wrist camera mounted near the end-effector. This two-view design is used consistently throughout rendering, AXIS pretraining, and LIBERO post-training.

During rendering, the simulator state is set explicitly at each timestep. The replay record includes robot joint position, joint velocity, joint position target, object position, object rotation, object linear velocity, and object angular velocity. The joint position target is retained because it supports downstream action supervision and consistency checks, while the object state fields are required to reconstruct task-relevant interactions without relying on a fresh dynamics rollout.

In the IsaacSim rendering path, physics stepping is disabled during replay. The replay state is therefore authoritative: the renderer refreshes cameras and sensors after the state is set, but it does not use a new physics integration step to determine the next state. This makes the data-generation process closer to visualizing a verified trajectory than to resimulating a controller. It also explains why table-height or scene-geometry randomization must be coupled with corresponding adjustments to replayed robot/object states and camera look-at targets. The randomized scene changes the visual and geometric context, but the replayed trajectory must remain physically placed on the randomized table.

Validation is performed at the demonstration level. A demonstration is rejected if required RGB videos or metadata are missing, if joint traces are empty, non-finite, all zero, length-mismatched, or too short for action-chunk training. The validated output contains the third-view RGB stream, wrist RGB stream, metadata, joint-position trace, joint-position-target trace, and episode boundary information. These elements are then packaged into training-ready records with third-view RGB, wrist RGB, state, and 7D LIBERO-style joint-position actions.

The default rendering backend uses IsaacSim / IsaacLab with ray-traced lighting. Depth is disabled by default, so the stable visual data consist of third-view RGB and wrist RGB. The third-view camera approximates a fixed external RGB sensor, while the wrist camera is mounted on the Franka hand and approximates an eye-in-hand sensor. Both camera pose randomization and full scene randomization are enabled.

\begin{table}[t]
\centering
\small
\caption{Default rendering, camera, and domain-randomization parameters.}
\label{tab:axis_render_camera_dr}
\setlength{\tabcolsep}{3pt}
\renewcommand{\arraystretch}{1.06}
\begin{tabularx}{\textwidth}{@{}p{0.27\textwidth}p{0.28\textwidth}X@{}}
\toprule
\textbf{Category} & \textbf{Parameter} & \textbf{Value} \\
\midrule
Rendering backend & Simulator & IsaacSim / IsaacLab \\
Rendering backend & Render mode & Ray-traced lighting \\
Rendering backend & Render resolution & $256 \times 256$ \\
Rendering backend & Camera data type & RGB \\
Rendering backend & Depth & Disabled by default \\
Robot & Robot model & Franka, fixed base, position control \\
Third-view camera & Position & $(1.21, 0.0, 0.885)$ \\
Third-view camera & Look-at target & $(0.3, 0.0, 0.0)$ \\
Third-view camera & Focal length & 24.0 \\
Third-view camera & RGB field of view & 70$^\circ$ horizontal, 43$^\circ$ vertical \\
Third-view camera & Focus distance & 400.0 \\
Third-view camera & Clipping range & $(0.05, 100000.0)$ \\
Wrist camera & Mount link & Franka hand \\
Wrist camera & Local position & $(0.16, 0.0, 0.01)$ \\
Wrist camera & Local quaternion & $(0.0, -0.38268343, 0.0, -0.92387953)$ \\
Wrist camera & RGB field of view & 87$^\circ$ horizontal, 58$^\circ$ vertical \\
Third-view camera randomization & Position delta & $\pm 0.10$ m along each axis \\
Third-view camera randomization & Roll/pitch/yaw delta & $\pm 8^\circ$ \\
Wrist camera randomization & Depth / lateral / vertical delta & $\pm 0.02$ m, $\pm 0.015$ m, $\pm 0.005$ m \\
Wrist camera randomization & Roll / pitch / yaw delta & $\pm 8^\circ$, $\pm 4^\circ$, $\pm 4^\circ$ \\
Scene randomization & Randomization level & Level 3 \\
Scene randomization & Scene mode & Mode 3, full USD scene \\
Lighting randomization & Main light intensity & 12,000--35,000 \\
Lighting randomization & Corner light intensity & 5,000--15,000 \\
Lighting randomization & Main color temperature & 2,800--6,500 K \\
Lighting randomization & Sphere color temperature & 2,500--6,000 K \\
Lighting randomization & Disk orientation delta & $\pm 15^\circ$ \\
Lighting randomization & Sphere position delta & $\pm(0.5, 0.5, 0.3)$ m \\
\bottomrule
\end{tabularx}
\end{table}

The randomization settings are broader than simple camera jitter. Scene mode 3 enables full USD scene composition, and level 3 includes scene, material, lighting, and camera variation. The intended effect is to increase visual diversity while keeping the underlying state/action supervision tied to verified trajectories. Because depth is disabled by default, most of the rendered data volume comes from RGB videos, metadata, and wrist RGB observations.

\section{$\pi_{0.5}$ Continual Pretraining}
\label{sup:pi05-pretraining}

AXIS pretraining uses verified episodes converted into a LIBERO-style visual imitation-learning format. The default snapshot follows the paper-wide AXIS release scale: 207 tasks and 50,129 episodes. Each episode pairs continuous robot state/action supervision with a third-view camera, a wrist camera, and episode-boundary information.

\begin{table}[H]
\centering
\small
\caption{AXIS-Datasets snapshot and default $\pi_{0.5}$ input dimensions.}
\label{tab:axis_pretrain_dataset}
\setlength{\tabcolsep}{4pt}
\renewcommand{\arraystretch}{1.08}
\begin{tabularx}{\textwidth}{@{}p{0.45\textwidth}X@{}}
\toprule
\textbf{Dataset / Model Parameter} & \textbf{Value} \\
\midrule
AXIS task count & 207 \\
AXIS episode count & 50,129 \\
Observation cameras & Third-view RGB and wrist RGB \\
Image resolution & $256 \times 256$ \\
State dimension & 8D robot state \\
Action dimension & 7D joint-position action \\
Action horizon & 10 \\
Model action dimension after padding & 32 \\
Model state dimension after padding & 32 \\
\bottomrule
\end{tabularx}
\end{table}

Training samples are formed as observation-action chunks within individual episodes. For each selected timestep, the sampler reads the current state and image observations, then constructs a 10-step action chunk from the same episode. If the timestep is close to an episode boundary, the action targets are clipped at the episode end. This prevents action supervision from crossing into the next demonstration and makes the action target semantically consistent with the current observation.
The training representation is intentionally simple: each sample contains robot state, a 10-step action target, third-view RGB, wrist RGB, and the episode boundary needed to keep action chunks within a single demonstration. This representation is then mapped into the LIBERO-compatible model input format.
Pretraining initializes from the official $\pi_{0.5}$ base JAX parameters and performs full-model finetuning. The model uses a PaliGemma Gemma-2B backbone and a Gemma-300M action expert. No LoRA adapter is used: the vision encoder, language backbone, and action expert are all trainable. $\pi_{0.5}$ mode is enabled, the action horizon is 10, and the tokenizer uses a maximum prompt length of 200 tokens.
The input view set is intentionally narrow and matches the rendered dataset: the third-view image provides the external scene view, and the wrist image provides eye-in-hand context. State remains a continuous model input rather than being discretized into the language prompt. This keeps language reserved for task instructions while preserving low-level robot state in the continuous observation pathway.

The transform stack first maps third-view and wrist observations into the LIBERO-style observation layout, standardizes image layout, and applies quantile normalization to state and actions. Images are pad-resized from $256 \times 256$ to $224 \times 224$. During model loss computation, visual augmentation further applies random crop to 95\%, rotation in $[-5^\circ, 5^\circ]$, and brightness/contrast/saturation jitter. Prompts are tokenized with the PaliGemma SentencePiece tokenizer. Finally, state is padded from 8D to 32D and action chunks are padded from 7D to 32D; inference slices the output back to the first seven action dimensions.

For $\pi_{0.5}$, state and action normalization use quantile statistics:
\[
x_{\mathrm{norm}} =
\frac{x-q_{01}}{q_{99}-q_{01}+10^{-6}} \times 2.0 - 1.0 .
\]
The statistics are computed over states and actions. Action statistics use the same 10-step horizon as training, including clipping at episode boundaries. Images are not normalized through these statistics; they are converted from uint8 image values into the model's image range during observation construction.

The pretraining loss is flow matching over action chunks, not one-step behavior cloning. Given normalized and padded actions with shape $[\mathrm{batch}, 10, 32]$, the training step samples Gaussian noise and a continuous timestep, forms a noisy action suffix, and trains the model to predict the velocity target:
\[
\epsilon \sim \mathcal{N}(0, 1), \qquad
t \sim \mathrm{Beta}(1.5, 1)\times 0.999 + 0.001,
\]
\[
x_t = t\epsilon + (1-t)a,\qquad
u_t = \epsilon-a .
\]
The model receives observation prefix tokens and noisy action suffix tokens, then predicts the action velocity for the final action-horizon positions. The loss is mean squared error between the predicted velocity and $u_t$, averaged over the action horizon and action dimensions. Since actions are padded to 32 dimensions, the unused dimensions become fixed zero targets unless explicit action-loss weights are introduced. Thus, the model is optimized over the full padded action interface even though only the first seven dimensions are returned at inference time.

\begin{table}[H]
\centering
\small
\caption{Default $\pi_{0.5}$ AXIS pretraining optimization settings.}
\label{tab:axis_pretrain_optimization}
\setlength{\tabcolsep}{4pt}
\renewcommand{\arraystretch}{1.08}
\begin{tabularx}{\textwidth}{@{}p{0.45\textwidth}X@{}}
\toprule
\textbf{Training Parameter} & \textbf{Value} \\
\midrule
Global batch size & 8 \\
Training hardware & 8$\times$A100 GPUs \\
Train steps & 100,000 \\
Optimizer & AdamW \\
AdamW $\beta_1$ & 0.9 \\
AdamW $\beta_2$ & 0.95 \\
AdamW $\epsilon$ & $10^{-8}$ \\
Weight decay & $10^{-10}$ \\
Gradient clipping & 1.0 global norm \\
Peak learning rate & $5\times10^{-5}$ \\
Warmup steps & 10,000 \\
Post-warmup learning rate & Effectively constant at $5\times10^{-5}$ \\
EMA decay & 0.999 \\
\bottomrule
\end{tabularx}
\end{table}

Training length is reported in optimization steps rather than epochs, because each sample is an observation-action chunk drawn from within an episode. Checkpoints contain training state, inference-facing parameters, and dataset normalization artifacts. When EMA is enabled, the inference-facing parameters use EMA-smoothed weights and are the intended source for downstream LIBERO post-training initialization.

\section{LIBERO Post-Training from AXIS-Pretrained Parameters}
\label{sup:libero-post-training}

LIBERO post-training starts from AXIS-pretrained $\pi_{0.5}$ inference parameters. Only model parameters are transferred. Optimizer state, training step, and EMA state are initialized for a new LIBERO run. This is different from resuming training: resume is only appropriate when continuing the same LIBERO run, because it restores the full training state rather than using the AXIS checkpoint as an initialization source.

The AXIS-to-LIBERO transfer also keeps dataset normalization separate. The AXIS checkpoint provides weights, but LIBERO post-training uses LIBERO-specific normalization statistics stored with the post-training checkpoint. This prevents the LIBERO policy from using AXIS state/action normalization at serving time. The AXIS-to-LIBERO setup uses the final AXIS full-pretraining parameters as initialization and trains the LIBERO stage for 30,000 steps.
LIBERO post-training uses the official LIBERO LeRobot dataset. The data provide third-view camera images, wrist camera images, 8D continuous policy state, task language prompts, and native LIBERO delta actions. Because LIBERO actions are already delta actions, no additional delta conversion is applied. This keeps the post-training visual input consistent with the AXIS-pretraining setup.

The post-training input representation remains aligned with $\pi_{0.5}$. The third-view camera and wrist camera provide visual observations; the state remains an 8D continuous vector before padding, and the action target is a 10-step chunk. The output transform returns the first seven action dimensions. LIBERO-specific normalization statistics are stored with the post-training checkpoint and are required for downstream policy loading.

\begin{table}[H]
\centering
\small
\caption{Default LIBERO post-training data, input, and optimization settings.}
\label{tab:libero_post_training}
\setlength{\tabcolsep}{4pt}
\renewcommand{\arraystretch}{1.08}
\begin{tabularx}{\textwidth}{@{}p{0.45\textwidth}X@{}}
\toprule
\textbf{LIBERO / Post-training Parameter} & \textbf{Value} \\
\midrule
Dataset & Official LIBERO LeRobot release \\
Expected episodes & 1,693 \\
Image shape & $[256, 256, 3]$ \\
State dimension & 8D LIBERO policy state \\
Action horizon & 10 \\
Action semantics & Native LIBERO delta action \\
Extra delta conversion & Disabled \\
Active cameras & Third-view camera and wrist camera \\
Train steps & 30,000 \\
Global batch size & 64 \\
Training and evaluation hardware & 8$\times$A100 GPUs \\
Optimizer & AdamW \\
Gradient clipping & 1.0 global norm \\
Peak learning rate & $5\times10^{-5}$ \\
Warmup steps & 10,000 \\
Post-warmup learning rate & Effectively constant at $5\times10^{-5}$ \\
EMA decay & 0.999 \\
\bottomrule
\end{tabularx}
\end{table}

The default 8$\times$A100 setting uses a global batch size of 64, which preserves approximately eight samples per GPU. The optimization setup mirrors the pretraining learning rate and EMA choices while changing the data distribution and initialization.

A LIBERO post-training checkpoint must include inference parameters and LIBERO-specific normalization artifacts. Loading only the model weights is insufficient for downstream policy construction because normalization statistics are required to build the trained policy correctly. The checkpoint therefore stores inference-facing model weights, training state for resuming the same run, and dataset-specific normalization statistics.

\section{Real-Time Inference and Real-World Rollouts}
\label{sup:real-time inference}

This section supplements the main paper's simulation-based LIBERO-Plus evaluations with qualitative real-world rollouts. We evaluate real-time inference on a Franka robot equipped with a gripper. For the qualitative rollouts, we selected two tasks: grasping and pick-and-place.

\begin{figure}
    \centering
    \includegraphics[width=1\linewidth]{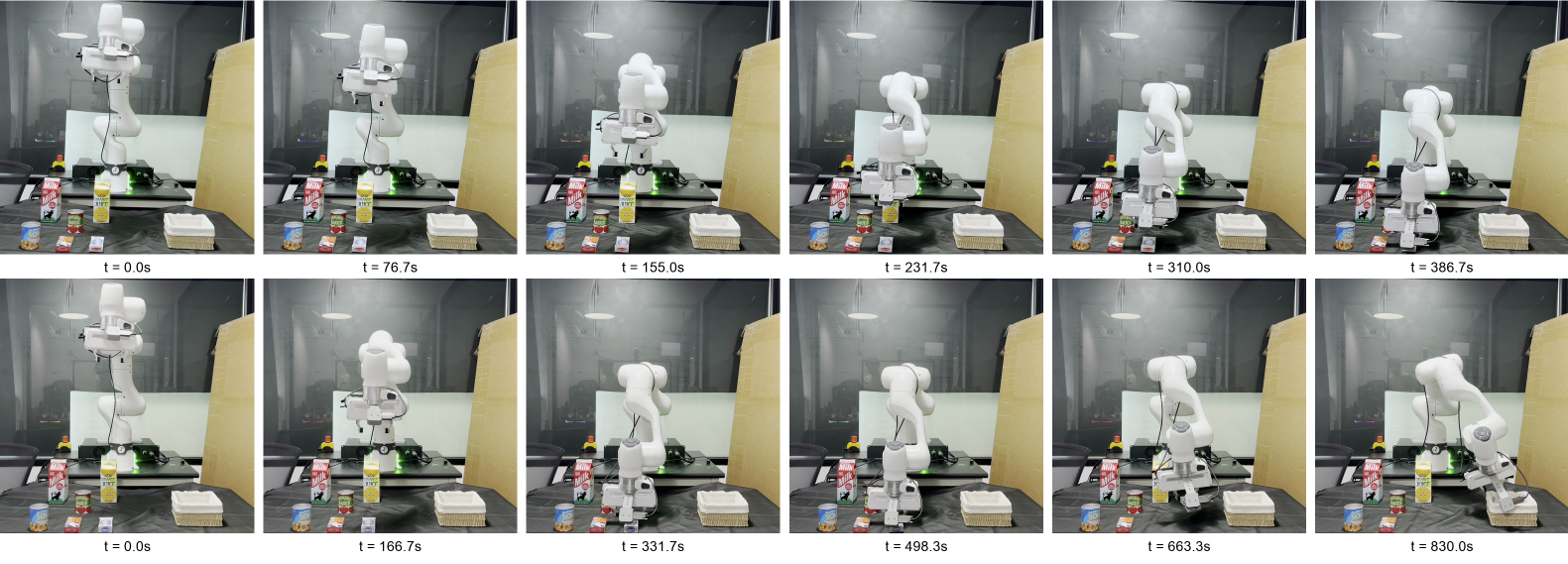}
    \caption{Real-world Franka rollouts. The first row shows a grasping task, and the second row shows a pick-and-place task. Timestamps indicate elapsed real execution time.}
    \label{fig:realtime_rollout}
\end{figure}

\begin{figure}
    \centering
    \includegraphics[width=\linewidth]{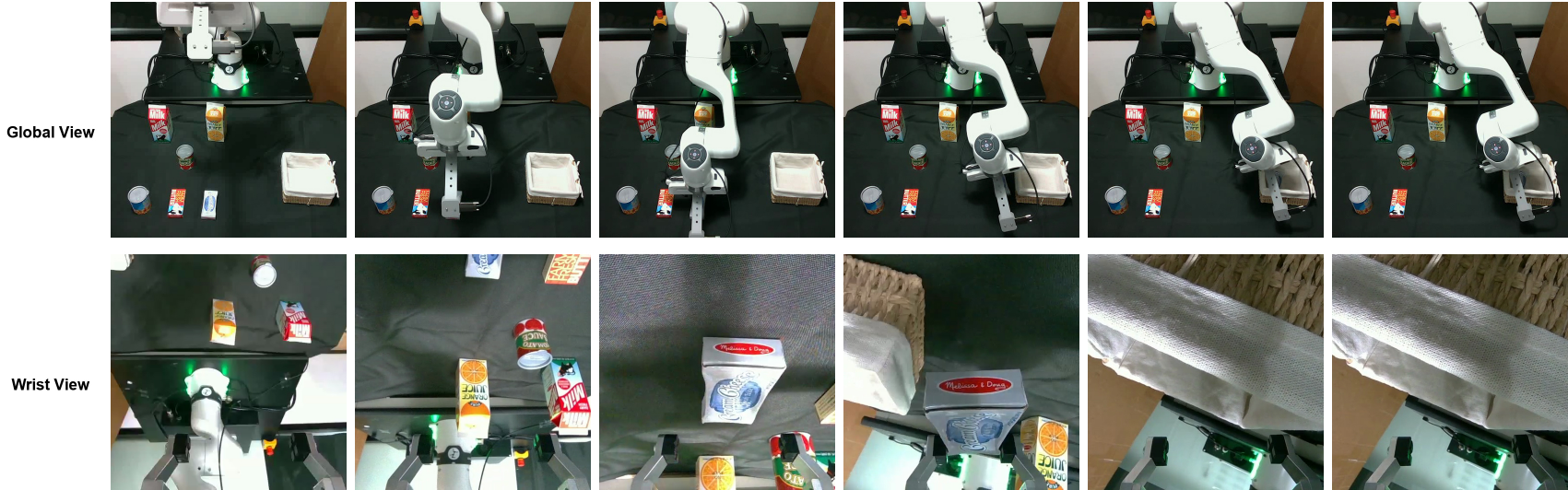}
    \caption{Real-time pick-and-place rollout. The two rows show the global and wrist views.}
    \label{fig:pick_place_rollout}
\end{figure}

During real-time inference, the policy receives RGB observations from a global camera and a wrist-mounted camera, together with the continuous robot state. The policy predicts a short-horizon action chunk, which is converted into executable commands for the Franka controller at deployment time. This setup uses the same types of visual and proprioceptive inputs as the training pipeline, while leaving calibration, timing, and low-level controller execution to the real-robot system.

\begin{figure}
    \centering
    \includegraphics[width=\linewidth]{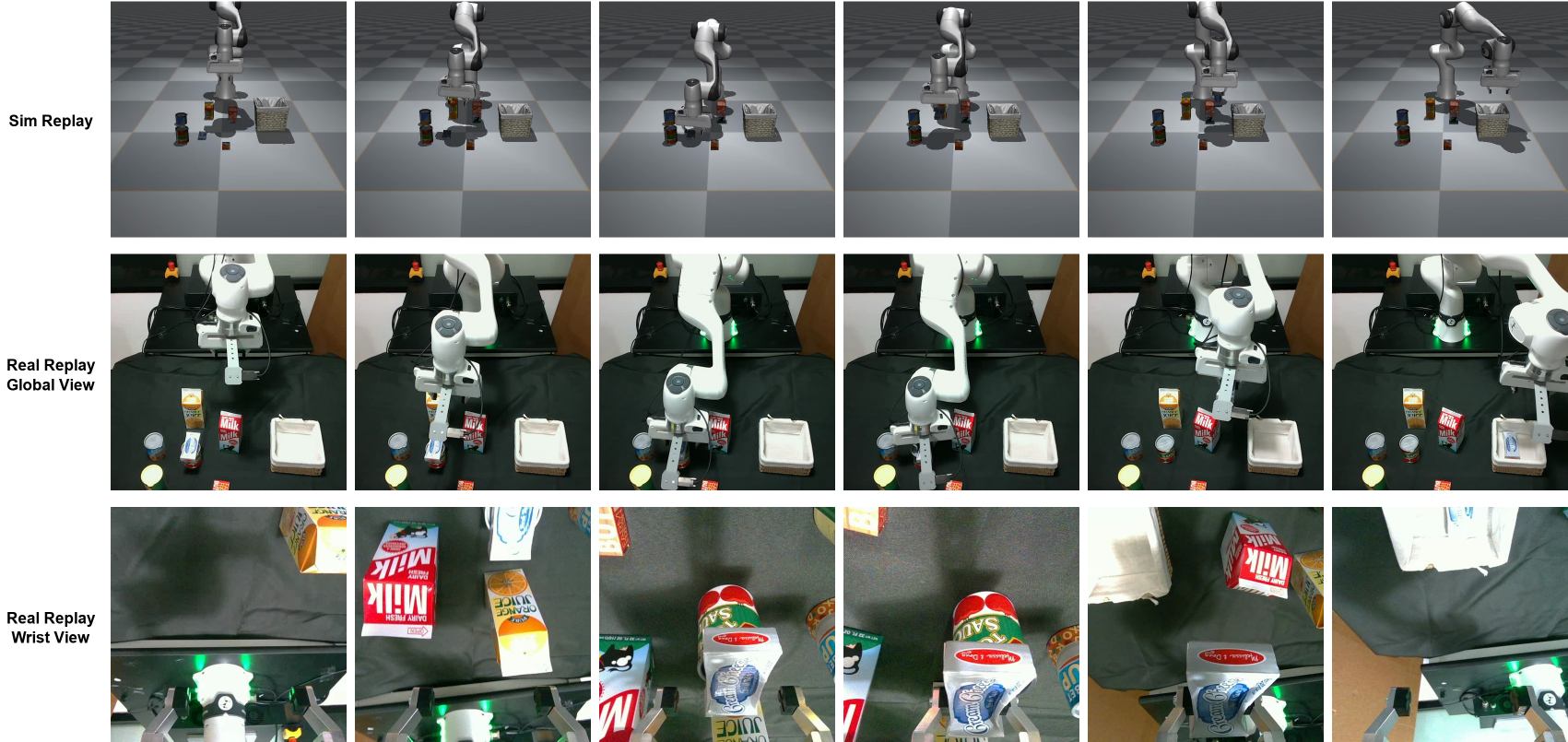}
    \caption{Pick-and-place replay. The three rows show the simulation replay, real-world global view, and real-world wrist view, respectively.}
    \label{fig:pick_place_replay}
\end{figure}

In addition to real-world keyframes, we also include a qualitative comparison between simulation replay and real-world execution. The simulation row visualizes the replayed task trajectory, while the real-world rows show the corresponding view observations. This comparison is intended to illustrate the observation-level correspondence between the replay pipeline and the real-robot setup. These examples are intended as qualitative evidence of data usability and policy executability.

\end{document}